\pdfoutput=1

\documentclass[11pt]{article}

\usepackage{ACL2023}

\usepackage{times}
\usepackage{latexsym}
\usepackage{booktabs}
\usepackage{array}
\usepackage{multirow}
\usepackage{soul} 
\usepackage{tcolorbox}
\usepackage{tikz}
\usepackage{float}
\usepackage{subcaption}
\usepackage{booktabs}
\usepackage{pgf-pie}
\usepackage{mdframed}
\usepackage{enumitem}
\newcommand{\squishlist}{
 \begin{list}{$\bullet$}
  { \setlength{\itemsep}{0pt}
     \setlength{\parsep}{3pt}
     \setlength{\topsep}{3pt}
     \setlength{\partopsep}{0pt}
     \setlength{\leftmargin}{1.5em}
     \setlength{\labelwidth}{1em}
     \setlength{\labelsep}{0.5em} } }

\newcommand{\squishlisttwo}{
 \begin{list}{$\bullet$}
  { \setlength{\itemsep}{0pt}
    \setlength{\parsep}{0pt}
    \setlength{opsep}{0pt}
    \setlength{\partopsep}{0pt}
    \setlength{\leftmargin}{2em}
    \setlength{\labelwidth}{1.5em}
    \setlength{\labelsep}{0.5em} } }

\newcommand{\squishend}{
  \end{list}  }
\usepackage[T1]{fontenc}

\usepackage[utf8]{inputenc}
\usepackage{amsmath}
\usepackage{microtype}

\usepackage{inconsolata}
\newcommand{\ourapproach}{{\sc Mace}}

\newcommand{\LL}[1]{#1}
\newcommand{\itb}[1]{\textit{\textcolor{blue}{#1}}} 

\newcommand{\rudra}[1]{{#1}}

\newcommand{\eat}[1]{} 

\title{A Multi-Agent Approach for Claim Verification from Tabular Data Documents}

\author{Rudra Ranajee Saha \\
  University of British \\ Columbia \\
  \texttt{rrs99@cs.ubc.ca} \\\And
  Laks V. S. Lakshmanan \\
  University of British \\ Columbia \\
  \texttt{laks@cs.ubc.ca} \\\And
  Raymond T. Ng \\
  University of British \\ Columbia \\
  \texttt{rng@cs.ubc.ca} \\}

\begin{document}
\maketitle
\begin{abstract}
We present a novel approach for claim verification from tabular data documents. Recent LLM-based approaches either employ complex pretraining/fine-tuning or decompose verification into subtasks, often lacking comprehensive explanations and generalizability. To address these limitations, we propose a Multi-Agentic framework for Claim vErification (\ourapproach{}) consisting of three specialized agents: Planner, Executor, and Verifier. Instead of elaborate fine-tuning, each agent employs a zero-shot Chain-of-Thought setup to perform its tasks. \rudra{\ourapproach{} produces interpretable verification traces, with the Planner generating explicit reasoning strategies, the Executor providing detailed computation steps, and the Verifier validating the logic.} Experiments demonstrate that \ourapproach{} achieves state-of-the-art (SOTA) performance on two datasets and performs on par with the best models on two others, while achieving 80--100\% of best performance with substantially smaller models: 27--92B parameters versus 235B. This combination of competitive performance, memory efficiency, and \rudra{transparent reasoning} highlights our framework's effectiveness.\footnote{Upon acceptance of the paper, we will release our codebase to the public.}
\end{abstract}

\section{Introduction}
Claim verification from tabular data documents is a critical task in Natural Language Processing, driven by the need to combat misinformation in domains where factual information is often stored in structured tables. Tables, embedded within documents such as scientific papers, financial reports, and corporate sustainability disclosures, serve as concise representations of complex information, making them pivotal for verifying claims about numerical, categorical, or relational data. \rudra{Documents may contain contextual paragraphs alongside tables.} \rudra{As organizations increasingly rely on tabular data to communicate key metrics, accurate claim verification against these sources is critical to ensure trust and accountability.}
\begin{table*}[h]
\centering
\small
\begin{tabular}{p{1.1cm}p{3.1cm}*{7}{p{1cm}}}
\toprule
\textcolor{blue}{Row No.} & \textbf{} & \textbf{2023} & \textbf{2022} & \textbf{2021} & \textbf{2020} & \textbf{2019} & \textbf{2018} & \textbf{2017} \\
\midrule
\textcolor{blue}{1} & \textbf{All Operations} & & & & & & & \\
\textcolor{blue}{2} & Water withdrawal(2) & 145,770 & 117,327 & 117,262 & 118,284 & 127,018 & 128,146 & 115,368 \\
... & ... & ... & ... & ... & ... & ... & ... \\
\textcolor{blue}{5} & Water reused/recycled & 168,358 & 134,131 & 138,812 & 157,641 & 148,914 & 174,688 & 176,563 \\
\textcolor{blue}{6} & Operational water use(4) & 315,784 & 250,449 & 256,074 & 275,925 & 275,931 & 302,835 & 291,931 \\
\midrule
\textcolor{blue}{7} & \textbf{Mining Operations} & & & & & & & \\
\textcolor{blue}{8} & Water withdrawal(2) & 72,645 & 47,701 & 45,222 & 47,739 & 51,954 & 60,003 & 44,225 \\
... & ... & ... & ... & ... & ... & ... & ... \\
\textcolor{blue}{12} & Water reused/recycled & 168,358 & 134,131 & 138,812 & 157,641 & 148,914 & 174,688 & 176,563 \\
\textcolor{blue}{13} & Operational water use(4) & 242,687 & 180,823 & 184,034 & 205,381 & 200,867 & 234,691 & 220,788 \\
\bottomrule
\end{tabular}
\caption{Water Metrics in Megalitres (ML)(1). (2)   Water withdrawal is water that enters the operational water system and is used to supply the operational water demands. ... (4)   Total water use is the sum of water withdrawals and water reused/recycled. 
\textcolor{blue}{Row No. column is added only for illustration purposes.}}
\label{tab:MineTabFact_table}
\end{table*} 
\rudra{Formally, the task takes a claim and evidence (tables/paragraphs) as input, outputting a verdict (``support,'' ``refute,'' or ``not enough info'') with reasoning. Throughout this paper, we use `\textit{evidence}' to refer to tables and paragraphs.} 
\rudra{This task spans \textit{closed-domain} \cite{Wu2024ProTrixBM, lu2024tart} (claims paired with specific tables) and \textit{open-domain} \cite{zhao2024findver, gu2022opentfv} (claims paired with document corpora, with models having to retrieve relevant tables from the corpora) settings.}
Table ~\ref{tab:MineTabFact_table} shows an example from a mining company's sustainability report. For claim \textit{`Reused/recycled water made up 55.82\% of all operational water use across all years'}, Table~\ref{tab:MineTabFact_table} provides supporting evidence (detailed analysis in Section~\ref{methodology}). Despite recent advancements, claim verification from tabular data documents faces challenges \rudra{that include} diverse table structures and the need for sophisticated reasoning to integrate tabular and textual evidence. \rudra{For instance, verifying this claim requires: identifying relevant rows based on scope (``All Operations'' vs ``Mining Operations''), understanding hierarchical structure (rows 2--6 under row 1), selecting data rows (5--6 for reused/recycled water and operational use), extracting values across years (2017--2023), and determining operation sequence (summation then percentage) to validate 55.82\%.} Additionally, open-domain retrieval, generating explainable outcomes, and generalizability across domains remain significant \rudra{challenges} \cite{gupta2022right}. To address these challenges, we propose (\ourapproach{}), a novel, prompt-based approach that avoids elaborate pre-training and fine-tuning. \ourapproach{} comprises \textbf{three specialized agents}: (i) \textit{Planner}--devises reasoning strategies, (ii) \textit{Executor}-- implements the plan and performs verification, and (iii) \textit{Verifier}--validates output for consistency and correctness. This design enables effective collaboration among agents while maintaining interpretability. \rudra{A key innovation is our feedback loop mechanism: the Executor can request plan revisions from the Planner when encountering ambiguities, and the Verifier can request corrections from the Executor when detecting inconsistencies, preventing error propagation.} Our framework achieves SOTA performance on two closed-domain datasets and matches the best models on two open-domain datasets, demonstrating its versatility in both claim verification settings. Smaller configurations (27--92B parameters) achieve 80--96\% of 235B model performance using 11.5--39.1\% memory, offering compelling performance-memory tradeoffs with step-by-step explainability \rudra{that details the selected rows, columns, cells, and mathematical operations performed}. 
\section{Related Work}
Prior works address claim verification through three main approaches: pre-training, fine-tuning, and prompt engineering.

\noindent 
\textbf{Pre-training and Fine-tuning Approaches.}
Recent advances include TableGPT2 \cite{su2024tablegpt2}, trained on 593.8K tables for schema and cell-level information, and TableRAG \cite{chen2024tablerag}, using retrieval-augmented generation for large table context limitations. Fine-tuning methods like TAT \cite{zhu2024tat} decompose reasoning into extraction, reasoning, and execution steps; ProTrix \cite{Wu2024ProTrixBM} employs a Plan-then-Reason framework for diverse table queries.
 \par
\noindent\textbf{Claim Verification from Tabular Data.}
FinDVer \cite{zhao2024findver} focuses on claim verification over financial documents,  and shows that even models like GPT-4o underperform human experts in numerical reasoning. TART \cite{lu2024tart} bridges program-based precision and chain-of-thought explainability by integrating LLMs with specialized tools, achieving SOTA performance on SciTab \cite{lu2023scitab} and PubHT \cite{akhtar2022pubhealthtab} benchmarks. For open-domain, OpenTab \cite{kong2024opentab} combines BM25 retrieval with \eat{multi-stage} SQL generation; GraphOTTER \cite{li2024graphotter} transforms tables into graph representations, effective for tables with merged cells and nested structures.

\noindent Despite this progress, several challenges persist. Many approaches rely on proprietary large models like GPT-3/4 \citep{zhao2024findver, lu2024tart}, requiring substantial resources and hindering reproducibility. Fine-tuned models excel on specific datasets but struggle to generalize across diverse table formats and domains, while resource-intensive methods limit practicality. Furthermore, unidirectional task decomposition techniques \citep{ye2023large,zhao2024tapera} are prone to error propagation, while SQL generation methods face difficulties with complex header hierarchies \citep{li2024graphotter}. In contrast, our \ourapproach{} framework incorporates feedback loops enabling iterative refinement, revision requests when detecting inconsistencies, mitigating error propagation.
\section{Methodology}\label{methodology}
\begin{figure*}
    \centering
    \includegraphics[width=0.85\textwidth]{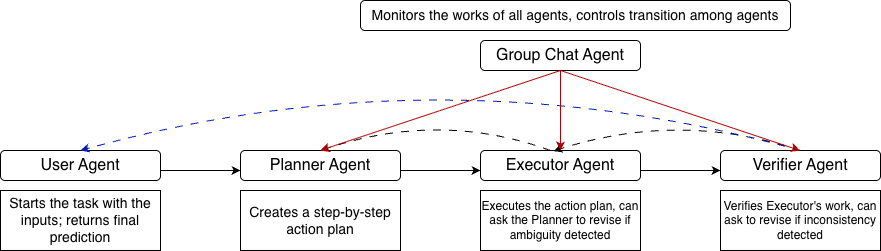}
    \caption{Overview of the \ourapproach{} framework. \rudra{Black solid arrows indicate the standard workflow sequence among agents. Black dashed arrows represent feedback loops; Red arrows show the Group Chat agent monitoring all agents. Blue dashed arrows indicate the final verdict return path from the Verifier to the User agent.}
    }
    \label{fig:mace_overview}
\end{figure*}
\ourapproach{} is implemented using the AutoGen library~\cite{wu2023autogen} and it functions as a structured, constrained group-chat system where agents collaborate in a predefined sequence to verify claims against tabular data documents (Figure~\ref{fig:mace_overview}).  It comprises four agents: User agent provides the claim and evidence; Planner agent designs a step-by-step verification strategy; Executor agent implements the plan and produces a provisional verdict; and Verifier agent evaluates the output for consistency and correctness. The group-chat operates through constrained speaker transitions managed by a custom selection function, following: User $\rightarrow$ Planner $\rightarrow$ Executor $\rightarrow$ Verifier $\rightarrow$ User. The framework incorporates a revision mechanism enabling the Executor to request plan revisions from the Planner, and the Verifier to request corrections from the Executor—a key advantage over unidirectional pipelines where errors propagate through subsequent stages. Each agent is configured with detailed system messages but operates in zero-shot mode. \rudra{The framework supports closed-domain (claims paired with specific tables) as well as open-domain (adding a retrieval module before Planner for retrieving relevant evidence from corpora) settings.}
\rudra{Throughout this section, all examples illustrate the verification process for the claim \textit{`Reused/recycled water made up 55.82\% of all operational water use across all years'} using data from Table ~\ref{tab:MineTabFact_table}.}

\begin{figure*}[t]
    \centering
    \begin{tikzpicture}
        \node [rectangle, draw=black, inner sep=5pt, text width=0.48\textwidth, align=left] at (0,0) {
            \textbf{Steps for the Executor:} These steps are provided by the Planner for the Executor to perform.\par
            \vspace{0.1cm}
            1. Identify relevant data rows - extract "Water reused/recycled" and "Operational water use(4)" for "All Operations" (2023--2017) (\itb{rows 5 and 6} of Table \ref{tab:MineTabFact_table}).\par
            \vspace{0.2cm}
            2. Compute yearly percentages and the overall percentage.\par
            \vspace{0.2cm}
            3. Compare the overall percentage to 55.82\%.\par
            \vspace{0.2cm}
            4. Return "support" if the computed percentage is 55.82\% (or very close, e.g., rounding to two decimal places), otherwise "refute."
        };
        \node [rectangle, draw=black, inner sep=5pt, text width=0.48\textwidth, align=left] at (0.51\textwidth,0) {
            \textbf{Results from the Executor agent:} Action plan performed by the Executor agent.\par
            \vspace{0.2cm}
            1. Extract "Water reused/recycled" for All Operations (2017-2023)- 2017: 176563 ML, ..., 2023: 168358 ML, Total Reused/Recycled = 176563 + ... + 168358 = 1,099,107 ML. (\itb{rows 5 and 6} of Table \ref{tab:MineTabFact_table}).\par
            2. Extract `Operational water use(4)' for All Operations (2017-2023)- ...  = 1,968,929 ML\par
            3. Compute overall percentage: $(1,999,107 / 1,968,929) \times 100 = 55.82\%$\par
            4. Compare with claimed percentage (55.82\%): the computed percentage (55.82\%) matches the claimed percentage exactly. Support.
        };
    \end{tikzpicture}
    \caption{Snippets of the action plan generated by the Planner agent (left) and execution results from the Executor agent (right). Text in \textit{italic} and colored \textcolor{blue}{blue} are added for illustration purposes only.}
    \label{fig:executor_steps}
\end{figure*}

\subsection{Planner Agent}
\label{sec:planner}

The Planner (Figure~\ref{fig:mace_overview}) receives the claim and evidence from the User agent, then devises an action plan for verification. It breaks down the verification task into explicit, actionable steps that the Executor can follow without ambiguity. The agent addresses several critical challenges: First, it pays attention to negation words such as "not", "no", "without", "does not", etc., which can drastically alter claim meaning and must be reflected in the verification strategy. Second, it resolves ambiguous descriptors (e.g., "significant," "outperform," "large margin") through quantitative thresholds based on context or domain norms. Third, it manages compound claims by identifying parts that need verification and determining dependencies between them. The Planner produces an \textbf{<explanation>} (identifying key claim components, relevant table columns/rows, and comparisons) and a \textbf{<plan>} (instructions for data extraction, computation, and comparison). \rudra{Figure \ref{fig:executor_steps}(left) shows an action plan snippet generated by the Planner (see Figure \ref{fig:action_plan from planner agent} for the complete plan), which correctly identifies the necessary rows, mathematical operations, and the verification threshold.}

\subsection{Executor Agent}
\label{sec:executor}

The Executor (Figure~\ref{fig:mace_overview}) receives the Planner's action plan and retrieves the original claim and evidence from group-chat history, then implements the plan with explicit reasoning traces through necessary computations. The agent executes each step while addressing several verification challenges. For verdict determination, it distinguishes between three outcomes: "support" when the table provides clear evidence backing the claim, "refute" when evidence contradicts the claim or the main component of a compound claim fails, and "not enough info" when required data is absent. The agent handles compound claims by treating the main point as decisive without forcing "refute" verdicts for auxiliary unverifiable details. It performs directional and numeric checks, attending to trend directions (increase vs decrease) and accepting approximate matches within a reasonable tolerance. For ambiguous terms, the Executor tests multiple reasonable interpretations and documents all results. When the instructions are unclear, the Executor outputs "revise" instead of a verdict, requesting clarification. The Executor produces \textbf{<explanation>} (execution trace showing data extraction, computations, and reasoning), a \textbf{final verdict} (a single line containing "support," "refute," or "not enough info"), or \textbf{"revise"} with explanation when the plan requires clarification. \rudra{Figure \ref{fig:executor_steps}(right) shows an execution trace snippet (see Figure \ref{fig:action_plan from executor agent} for the complete trace), which accurately extracts the relevant cells from Table \ref{tab:MineTabFact_table}, performs the calculations correctly, and arrives at the appropriate verdict.}

\subsection{Verifier Agent}
\label{sec:verifier}
\rudra{The Verifier agent (Figure~\ref{fig:mace_overview}) receives the Executor's explanation and provisional verdict, then audits the output for correctness and consistency before finalization. It acts as a quality control mechanism, catching errors before final verdict submission and preventing invalid outputs from reaching end users—unlike pipelines processing data without validation. The agent checks logical coherence between explanation and verdict, ensuring reasoning faithfully supports the decision, then follows two pathways: approve if reasoning and decision are sound, or output "revise" if errors, omissions, or inconsistencies exist. The system terminates when the Verifier issues a final verdict. Output includes \textbf{<explanation>} (validation reasoning examining consistency and correctness), \textbf{final verdict} ("support," "refute," or "not enough info" if approved), or \textbf{"revise"} with correction requests.}
\section{Experiment}\label{experiments}
We evaluate \ourapproach{} on two key questions: (1) How does \ourapproach{} perform compared to single-model approaches across model sizes, agent configurations, and resource (memory, runtime)? (2) How does \ourapproach{} compare against SOTA baselines in both closed- and open-domain settings? We organize this section as follows: Section~\ref{evaluation-datasets} describes datasets, ~\ref{implementation-details} details implementation, ~\ref{performance-analysis-of-multi-agent-configuration} analyzes \ourapproach{} configurations and resource efficiency, and ~\ref{comparison-against-baselines} compares against baselines.

\begin{table*}[h]
\centering
\small
\begin{tabular}{llcccp{5cm}}
\toprule
\textbf{Setting} & \textbf{Dataset} & \textbf{Tables} & \textbf{Claims} & \textbf{Labels} & \textbf{Key Characteristics} \\
\midrule
\multirow{2}{*}{Closed} 
& SciTab \cite{lu2023scitab} & 213 & 1,224 & 3-way & Claims from scientific publications \\
& SEM-TAB-FACTS \cite{wang2021semeval} & 2,961 & 653 & 3-way & Scientific tables, complex reasoning \\
\midrule
\multirow{4}{*}{Open} 
& \rudra{FinDVer \cite{zhao2024findver}} & & & \multirow{4}{*}{2-way} & \multirow{4}{*}{\rudra{Long, hybrid-content financial documents}} \\
& \hspace{3mm} \rudra{- Testmini} & \rudra{517} & \rudra{700} & & \\
& \hspace{3mm} \rudra{- Test} & \rudra{1,262} & \rudra{1,700} & & \\
& SciTab-OD & 213 & 868 & 2-way & Open-domain version of SciTab \\
\bottomrule
\end{tabular}
\caption{Evaluation datasets for claim verification. Labels: 3-way (support/refute/not enough info), 2-way (support/refute). \rudra{\textbf{For brevity, we refer to SEM-TAB-FACTS, Testmini, and Test as SemTab, TM, and T.}}}
\label{tab:datasets}
\end{table*}

\subsection{Evaluation Datasets}
\label{evaluation-datasets}
We evaluate \ourapproach{} across two verification settings: closed-domain and open-domain. Table~\ref{tab:datasets} summarizes the datasets.

\noindent\textbf{SciTab-OD Dataset Creation.} We created an open-domain version of the SciTab dataset (SciTab-OD) using retrieval-augmented generation. Unlike the SciTab dataset with one-to-one claim-table mapping, we assume no pairing. Following \cite{zhao2024findver}, we use only support/refute claims, yielding 213 tables and 868 claims.

\noindent\textbf{Retrieval Mechanism.} Since retrieval is not our primary focus, we adopt existing strategies. 
For FinDVer, \citet{zhao2024findver} compared three retrievers: BM25 \cite{robertson1995okapi}, Contriever \cite{izacard2021unsupervised}, and OpenAI's text-embedding-3 across retrieval sizes $k = 3, 5, 10$, finding text-embedding-3 with $k=10$ optimal. We adopt this configuration, achieving 67.91\% and 69.53\% recall for evidence retrieval on Testmini and Test sets. For SciTab-OD, we retrieve top-2 tables using text-embedding-3, achieving 68.43\% recall. 

\noindent\textbf{Evaluation Metrics.} To ensure fair comparison with prior works, we adopt \LL{the original metrics published with the datasets}: Macro F1 for SciTab and SciTab-OD, Micro F1 for SemTab, and Accuracy for FinDVer, enabling direct baseline comparison. 

\subsection{Implementation Details}
\label{implementation-details}

We evaluate \ourapproach{} using 4 Together.ai models at different scales: small (<70B: Mistral-7B, LLama-8B), medium (70-200B: Qwen-72B), large (>\LL{200B}: Qwen-235B), assigned in different combinations to agents. GPT-20B serves as the Verifier in some configurations to evaluate independent verification. For brevity, we refer to Mistral-7B as Mt-7B, LLama-8B as Ll-8B, Qwen-72B as Qw-72B, and Qwen-235B as Qw-235B. Each agent uses a 2000-second timeout, 3 retry attempts, temperature 0 for deterministic outputs, and 1500 maximum output tokens. Maximum conversation rounds is set to 8.
We handle 2-class (open-domain) and 3-class (closed-domain) classification by adjusting agent prompts. For baselines, we use zero-shot prompts from \citet{lu2023scitab} and \citet{zhao2024findver} for closed and open-domain, respectively. System messages are in Appendix Figures \ref{fig:planner_agent_overview_closed}--\ref{fig:user_agent_overview_closed} (closed-domain) and \ref{fig:planner_agent_overview}--\ref{fig:user_agent_overview} (open-domain). For closed-domain, tables are converted to HTML format. For open-domain, we use the retrieval results directly as provided by the respective retrieval systems.

\begin{table*}[t]
\centering
\small
\resizebox{\textwidth}{!}{
\begin{tabular}{ll|cc|ccc}
\toprule
\multicolumn{2}{c|}{\textbf{Model Configurations}} & \multicolumn{2}{c|}{\textbf{Closed-Domain}} & \multicolumn{3}{c}{\textbf{Open-Domain}} \\
\cmidrule(lr){1-2} \cmidrule(lr){3-4} \cmidrule(lr){5-7}
\textbf{Model} & \textbf{Configuration} & \textbf{SciTab} & \textbf{SemTab} & \textbf{FinDVer-TM} & \textbf{FinDVer-T} & \textbf{SciTab-OD} \\
 &  & \textbf{(Macro F1)} & \textbf{(Micro F1)} & \textbf{(Accuracy)} & \textbf{(Accuracy)} & \textbf{(Macro F1)} \\
\midrule
\multirow{5}{*}{Mt-7B} 
& w COT & \underline{0.49} & \underline{0.48} & \underline{0.56} & \underline{0.55} & \textbf{0.65} \\
& wo COT & 0.40 & 0.41 & 0.52 & 0.53 & 0.58\\
& \ourapproach{} (P$_m$ + E$_m$) & 0.34 & 0.26 & 0.52 & 0.51 & 0.50 \\
& \ourapproach{} (P$_m$ + E$_m$ + V$_m$) & 0.23 & 0.27 & 0.51 & 0.50 & 0.52\\
& \ourapproach{} (P$_m$ + E$_m$ + V$_{m'}$) & \textbf{0.58} & \textbf{0.74} & \textbf{0.64} & \textbf{0.65} & \underline{0.60}\\
\midrule
\multirow{5}{*}{Ll-8B} 
& w COT & \underline{0.53} & \underline{0.76} & \textbf{0.71} & \underline{0.70} &  0.56\\
& wo COT & 0.44 & 0.56 & 0.63 & 0.63 & 0.59\\
& \ourapproach{} (P$_m$ + E$_m$) & 0.46 & 0.58 & 0.63 & 0.62 & \underline{0.60}\\
& \ourapproach{} (P$_m$ + E$_m$ + V$_m$) & 0.45 & 0.60 & 0.62 & 0.61 & 0.53\\
& \ourapproach{} (P$_m$ + E$_m$ + V$_{m'}$) & \textbf{0.65} & \textbf{0.86} & \underline{0.68} & \textbf{0.71} & \textbf{0.66}\\
\midrule
\multirow{5}{*}{Qw-72B} 
& w COT & 0.62 & \underline{0.87} & \textbf{0.75} & \textbf{0.77} & \textbf{0.72}\\
& wo COT & \textbf{0.68} & 0.80 & \underline{0.73} & 0.72 & 0.69\\
& \ourapproach{} (P$_m$ + E$_m$) & 0.63 & \underline{0.87} & \textbf{0.75} &  0.73 & 0.70\\
& \ourapproach{} (P$_m$ + E$_m$ + V$_m$) & 0.64 & \underline{0.87} & \textbf{0.75} & \underline{0.74} & \underline{0.71}\\
& \ourapproach{} (P$_m$ + E$_m$ + V$_{m'}$) & \underline{0.67} & \textbf{0.89} & 0.72 & 0.72 & 0.66\\
\midrule
\multirow{5}{*}{Qw-235B} 
& w COT & \textbf{0.73} & \underline{0.89} & \textbf{0.76} & \underline{0.74} & 0.74\\
& wo COT & 0.70 & 0.84 & 0.72 & 0.73 & 0.68\\
& \ourapproach{} (P$_m$ + E$_m$) & \underline{0.71} & \textbf{0.90} & \textbf{0.76} & \textbf{0.76} & \underline{0.75}\\
& \ourapproach{} (P$_m$ + E$_m$ + V$_m$) & \underline{0.71} & \underline{0.89} & \underline{0.75} & \textbf{0.76} & \textbf{0.76}\\
& \ourapproach{} (P$_m$ + E$_m$ + V$_{m'}$) & \underline{0.71} & \underline{0.89} & 0.71 & 0.73 & 0.73\\
\bottomrule
\end{tabular}
}
\caption{Performance comparison of different model configurations across closed-domain and open-domain datasets. P: Planner, E: Executor, V: Verifier, $m'$: GPT-20B. For each dataset, \textbf{bold} indicates best performance and \underline{underline} indicates second-best performance across each model.}
\label{tab:model_configs}
\end{table*}

\subsection{Performance Analysis of Multi-Agent Configuration}
\label{performance-analysis-of-multi-agent-configuration}

We assess \ourapproach{} across five distinct setups for each model: single-model inference (i) with Chain-of-Thought prompting (w COT), and (ii) without COT (wo COT), and three \ourapproach{} variants including: (iii) two-agent configuration with Planner and Executor both using the same model $m$ (\ourapproach{} P$_m$ + E$_m$), where we remove the Verifier module and the Executor returns the final verdict directly to the User Agent; three-agent configuration where (iv) all agents use the same model $m$ (\ourapproach{} P$_m$ + E$_m$ + V$_m$); and (v) Planner and Executor use model $m$ while Verifier uses GPT-20B, denoted as $m'$ (\ourapproach{} P$_m$ + E$_m$ + V$_{m'}$). We introduce the fifth variant to investigate whether \ourapproach{} benefits more from using the same model across all agents or from employing a different model as the Verifier. Table \ref{tab:model_configs} presents the results across two closed and two open-domain datasets. Following \citet{zhao2024findver}, we employ GPT-4o to extract the final verdict from model outputs in the w COT and wo COT setups. In contrast, no \ourapproach{} variants require GPT-4o post-processing, as the structured agent framework ensures explicit verdict formatting. 

\subsubsection{Closed-Domain Performance}
\label{closed-domain-performance}

\rudra{\ourapproach{} demonstrates \eat{different benefits across model sizes with} remarkable gains for smaller models: Mt-7B improves from 0.49 to 0.58 on SciTab (18.4\%) and from 0.48 to 0.74 on SemTab (54.2\%), while Ll-8B improves from 0.53 to 0.65 on SciTab (22.6\%) and from 0.76 to 0.86 on SemTab (13.2\%). Medium and large models show more modest improvements: Qw-72B achieves 0.67 on SciTab (slightly below wo COT) and 0.89 on SemTab (2.30\% improvement), while Qw-235B reaches 0.71 on SciTab (vs 0.73 with COT) and 0.90 on SemTab (vs 0.89 with COT). This reveals that structured collaboration compensates for individual limitations in smaller models, allowing them to approach larger model capabilities, while larger models with strong reasoning see diminishing returns from coordination overhead.}



\subsubsection{Open-Domain Performance}
\label{sec:open-domain-performance}

\rudra{For smaller models, \ourapproach{} with independent verifier ($P_m + E_m + V_{m'}$) provides substantial benefits: Mt-7B improves from 0.56 to 0.64-0.65 on FinDVer (14.3-18.2\%) though it declines to 0.60 on SciTab-OD (vs 0.65 w COT), while Ll-8B achieves 0.68-0.71 on FinDVer and 0.66 on SciTab-OD (17.9\% improvement over 0.56 w COT). \rudra{Notably, \ourapproach{} configurations without independent verifier underperform significantly: on SciTab-OD, Mt-7B achieves only 0.50-0.52 (vs 0.60 with independent verifier), and on FinDVer-TM, Ll-8B reaches only 0.63 (vs 0.68 with independent verifier)}. For medium and large models, patterns shift: Qw-72B performs best with simple w COT on both FinDVer variants (0.75-0.77 vs \ourapproach{}'s 0.72-0.75) and SciTab-OD (0.72 vs 0.71), with the GPT-20B verifier configuration achieving only 0.66. Qw-235B with two-agent ($P_m + E_m$) achieving 0.76 on FinDVer and same-model three-agent ($P_m + E_m + V_m$) reaching 0.76 on SciTab-OD, while the GPT-20B verifier variant consistently underperforms (0.71-0.73). These results demonstrate that verifier selection should be adapted to base model capabilities: the GPT-20B verifier provides the most benefit for small models but for medium-large models, maintaining reasoning consistency within the same model family is more important than leveraging an independent verifier.}

\begin{table*}[t]
\centering
\small
\begin{minipage}[t]{0.55\textwidth}
\centering
\begin{tabular}{lcccc}
\toprule
Config. & Total & Memory & \multicolumn{2}{c}{Perf. vs 235B} \\
 & Params & vs 235B & SciTab & SemTab \\
\midrule
Qw-235B (baseline) & 235B & 100\% & 100\% & 100\% \\
\midrule
\multicolumn{5}{l}{\textit{Same-model verifier:}} \\
Qw-72B & 72B & 30.6\% & 87.7\% & 97.8\% \\
\midrule
\multicolumn{5}{l}{\textit{Independent verifier:}} \\
Mt-7B & 27B & 11.5\% & 79.5\% & 83.1\% \\
Ll-8B & 28B & 11.9\% & 89.0\% & 96.6\% \\
Qw-72B & 92B & 39.1\% & 91.8\% & 100\% \\
\bottomrule
\end{tabular}
\caption{Memory efficiency analysis for closed-domain datasets. All performance (Perf.) percentages are relative to Qw-235B w COT (0.73 on SciTab, 0.89 on SemTab).}
\label{tab:memory_efficiency}
\end{minipage}%
\hspace{0.2cm}
\begin{minipage}[t]{0.42\textwidth}
\centering
\scriptsize
\raisebox{4.2\normalbaselineskip}{%
\begin{tabular}{lcc}
\toprule
Config. & SciTab & SciTab-OD \\
\midrule
Mt-7B w COT & 4 & 12 \\
Mt-7B \ourapproach{} & 110 & 141 \\
\midrule
Qw-72B w COT & 23 & 62 \\
Qw-72B MACE & 123 & 106 \\
\bottomrule
\end{tabular}%
}
\vspace{0.15cm}

\begin{tabular}{lrr}
\toprule
Config. & SciTab & SemTab\\
\midrule
Qw-235B w COT & 55.07 & 42.13 \\
Ll-8B (\ourapproach{}) & 121.83 & 114.25\\
\midrule
Runtime Ratio & 2.21$\times$ & 2.71$\times$\\
\bottomrule
\end{tabular}
\caption{Runtime (minutes) comparison for 300 claims; (top) same-model and (bottom) memory-efficient vs. large model (Qw-235B).}
\label{tab:runtime_analysis}
\end{minipage}
\end{table*}

\subsubsection{Memory Efficiency Analysis}
\label{memory-efficiency-analysis}

\rudra{Memory footprint depends on configuration. Same-model ($P_m + E_m + V_m$) requires one instance (e.g., Qw-72B: 72B parameters). Independent verifier ($P_m + E_m + V_{m'}$) requires both (e.g., Mt-7B + GPT-20B = 27B).} Table~\ref{tab:memory_efficiency} compares these configurations against Qw-235B baseline and shows that smaller configurations achieve compelling performance with dramatically reduced memory. Configurations with 27--92B parameters reach \rudra{79.5--100}\% of Qw-235B's performance while consuming only 11.5--39.1\% of the memory. Notably, despite being nearly 30$\times$ smaller than Qw-235B, Ll-8B with \ourapproach{} achieves 89.0\% of the large model's performance on SciTab and 95.6\% on SemTab, while Mt-7B reaches 79.5\% and 83.1\% respectively. \rudra{On open-domain datasets, similar patterns emerge: Ll-8B (28B parameters) reaches 86.8--93.4\% of Qw-235B's performance on SciTab-OD and FinDVer while consuming only 11.9\% of the memory.} \rudra{These findings democratize high-quality claim verification, enabling deployment on hardware that is unable to support large models and allowing multiple simultaneous instances for increased throughput.}

\subsubsection{Runtime Analysis}
\label{runtime-analysis}

\rudra{While \ourapproach{} achieves strong performance with reduced memory, practical deployment requires \LL{analysis of}  execution time. Multi-agent coordination introduces overhead compared to a single-model inference. We investigate two questions for 300 claims: (1) How does \ourapproach{} runtime compare to single-model COT? (2) How does memory-efficient \ourapproach{} compare to the best-performing model?}

\noindent\textbf{Same-Model Comparison.} 
Table~\ref{tab:runtime_analysis} (top) compares single-model COT against \ourapproach{} with independent verifier ($P_m + E_m + V_{m'}$). For Mt-7B, \ourapproach{} takes 27.5$\times$ longer on SciTab (110 vs. 4 minutes) and 11.8$\times$ longer on SciTab-OD (141 vs. 12 minutes). This overhead reflects the cost of sequential agent processing with potential revision rounds. For Qw-72B, the overhead is lower: 5.3$\times$ on SciTab (123 vs. 23 minutes) and 1.7$\times$ on SciTab-OD (106 vs. 62 minutes). The performance gains justify these overheads. On SciTab, Mt-7B \ourapproach{} achieves 0.58 Macro F1 versus 0.49 with COT (18.4\% improvement). Qw-72B \ourapproach{} reaches 0.67 versus 0.62 with COT (8.1\% improvement). For scenarios prioritizing accuracy over latency, these tradeoffs are compelling.

\noindent\textbf{Memory-Efficient Configuration versus Large Model.} We compare Ll-8B \ourapproach{} ($P_m + E_m + V_{m'}$) against Qw-235B with COT.
Table~\ref{tab:runtime_analysis}(bottom) shows runtime across \rudra{two} closed-domain datasets. \rudra{The comparison reveals} 2.21--\rudra{2.71}$\times$ longer execution time than Qw-235B COT, averaging \rudra{2.46}$\times$ overhead. \rudra{Despite this overhead, the memory-efficient configuration} achieves 0.86 \rudra{on SemTab}, reaching \rudra{95.6}\% of Qw-235B's performance \rudra{(0.90)} with 28B versus 235B parameters. With less than \rudra{2.8}$\times$ runtime increase, the system reaches within \rudra{4.4}\% performance of the larger model, using only 11\% memory, enabling competitive verification in memory-constrained environments where deploying large models is infeasible.

\subsection{Comparison against Baselines}
\label{comparison-against-baselines}

We compare \ourapproach{} against SOTA baselines across closed and open-domain datasets, including fine-tuning, pipeline-based, ensemble systems, and LLMs (baseline details are in Appendix \ref{appendix:baseline-details}).

\subsubsection{Closed-Domain Datasets}
\label{performance-comparison-for-closed-domain-datasets}

\begin{table}[h]
\centering
\small
\begin{tabular}{|p{4.8cm}|c|}
\hline
\textbf{Method} & \textbf{Macro F1} \\ \hline
PASTA $^+$ & 0.33 \\ 
ProTrix $^{++}$ & 0.43 \\ 
TART (GPT-4)$^*$ & 0.64 \\ \hline
Alpaca-7B$^\dagger$ & 0.29 \\
Vicuna (13B)$^\dagger$ & 0.35 \\
Ll-13B$^\dagger$ & 0.33 \\ \hline
InstructGPT$^\dagger$ & 0.42 \\
InstructGPT + COT$^\dagger$ & 0.43 \\
GPT-4$^\dagger$ & 0.65 \\
GPT-4 + COT$^\dagger$ & 0.63 \\ \hline
\ourapproach{} (Mt-7B, $P_m + E_m + V_{m'}$) & 0.58 \\
\ourapproach{} (Ll-8B, $P_m + E_m + V_{m'}$) & 0.65 \\
\ourapproach{} (Qw-72B, $P_m + E_m + V_{m'}$) & \underline{0.67} \\
\ourapproach{} (Qw-235B, $P_m + E_m + V_{m'}$) & \textbf{0.71} \\
\hline
\end{tabular}
\caption{Macro F1 on SciTab dataset. Models marked with $^\dagger$, $^*$, and $^{++}$ are sourced from \citet{lu2023scitab}, \citet{lu2024tart} and \citet{Wu2024ProTrixBM} respectively; PASTA $^+$ is from executing \citet{gu2022pasta}.}
\label{tab:scitab_results}
\end{table}

\textbf{SciTab Dataset.} Table~\ref{tab:scitab_results} compares \ourapproach{} against pipeline-based approaches (PASTA, ProTrix, TART) and various LLMs.
\ourapproach{} achieves SOTA performance (0.71 Macro F1) without pre-training or fine-tuning, substantially outperforming all pipeline approaches, open-source models (Vicuna-13B) and closed-source models (GPT-4). Even smaller \ourapproach{} configurations demonstrate strong performance: Ll-8B matches GPT-4 at 0.65, while Mt-7B achieves 0.58, outperforming InstructGPT + COT (0.43).

\begin{table}[h]
\centering
\small
\begin{tabular}{|p{4.8cm}|c|}
\hline
\textbf{Method} & \textbf{Micro F1} \\
\hline
sattiy$^\dagger$ & 0.77 \\
RyanStark$^\dagger$ & 0.82 \\
THiFly-Queen$^\dagger$ & 0.84 \\
King001$^\dagger$ & 0.85 \\
\hline
Volta$^*$ & 0.74 \\
Tapas$^*$ & 0.75 \\
Tapex$^*$ & 0.76 \\
LKA$^*$ & 0.79 \\
DeBERTaV3$^*$ & 0.79 \\
PASTA$^*$ & 0.84 \\
\hline
\ourapproach{} (Mt-7B, $P_m + E_m + V_{m'}$) & 0.74 \\
\ourapproach{} (Ll-8B, $P_m + E_m + V_{m'}$) & 0.86 \\
\ourapproach{} (Qw-72B, $P_m + E_m + V_{m'}$) & \underline{0.89} \\
\ourapproach{} (Qw-235B, $P_m + E_m$) & \textbf{0.90} \\
\hline
\end{tabular}
\caption{Micro F1 on SemTab dataset. Models marked with $^\dagger$ are the best performing according to \citet{wang2021semeval}; those with $^*$ are sourced from \citet{gu2022pasta}.}
\label{tab:semtabfact_results}
\end{table}

\noindent\textbf{SemTab Dataset.} Table~\ref{tab:semtabfact_results} compares \ourapproach{} against \rudra{\href{https://aclanthology.org/2021.semeval-1.39/}{SemEval-2021 Task 9 competition winners}} and fine-tuned models. 
Mt-7B \ourapproach{} achieves 0.74, already competitive with many fine-tuned baselines despite being a smaller model. Ll-8B \ourapproach{} reaches 0.86, surpassing the previous SOTA King001 by 1.18\%. Qw-72B \ourapproach{} achieves 0.89, outperforming all \LL{competing} baselines. The largest configuration, Qw-235B \ourapproach{} achieves the best overall performance at 0.90.

\subsubsection{Open Domain Datasets}
\label{performance-comparison-for-open-domain-datasets}

\begin{table}[h]
\centering
\small
\begin{tabular}{|p{4.2cm}|c|c|}
\hline
 & \multicolumn{2}{c|}{\textbf{Accuracy}} \\
\cline{2-3}
\textbf{Method} & \textbf{TM} & \textbf{T} \\
\hline
DeepSeek-V2-Lite & 0.60 & 0.58 \\
Qwen-2.5 & 0.72 & 0.70 \\
Llama-3.1 70B & \underline{0.75} & \underline{0.75} \\
Qwen-2.5 72B & \textbf{0.76} & \underline{0.75} \\
Mistral-Large 123B & \underline{0.75} & \textbf{0.76} \\
\hline
Claude-3.5-Sonnet & 0.73 & 0.70 \\
Gemini-1.5-Pro & 0.71 & 0.73 \\
GPT-4o & \underline{0.75} & \textbf{0.76} \\
\hline
\ourapproach{} (Mt-7B, $P_m + E_m + V_{m'}$) & 0.64 & 0.65 \\
\ourapproach{} (Ll-8B, $P_m + E_m + V_{m'}$) & 0.68 & 0.71 \\
\ourapproach{} (Qw-72B, $P_m + E_m + V_{m}$) & \underline{0.75} & 0.74 \\
\ourapproach{} (Qw-235B, $P_m + E_m$) & \textbf{0.76} & \textbf{0.76} \\
\hline
\end{tabular}
\caption{Accuracy on FinDVer dataset. All baseline models are sourced from \citet{zhao2024findver}.}
\label{tab:findver_results}
\end{table}

\textbf{FinDVer Dataset.} Table~\ref{tab:findver_results} compares \ourapproach{} against open and closed-source models on FinDVer.
\rudra{\ourapproach{} achieves SOTA performance on both TM and T splits with Qw-235B ($P_m + E_m$) reaching 0.76 accuracy, matching the best baseline results. Qw-72B ($P_m + E_m + V_m$) achieves 0.75 on TM, matching the performance of larger baselines (Mistral-Large 123B, GPT-4o). Even smaller configurations show competitive results: Mt-7B ($P_m + E_m + V_{m'}$) reaches 0.64-0.65, outperforming DeepSeek-V2-Lite (0.60, 0.58), while Ll-8B ($P_m + E_m + V_{m'}$) achieves 0.68-0.71, approaching the performance of Qwen-2.5 (0.72, 0.70).}

\begin{table}[h]
\centering
\small
\begin{tabular}{|p{4.4cm}|c|c|}
\hline
 & \multicolumn{2}{c|}{\textbf{Macro F1}} \\
\cline{2-3}
\textbf{Method} & \textbf{w COT} & \textbf{wo COT} \\
\hline
Llama-3B & 0.57 & 0.37 \\
Mistral-7B & 0.65 & 0.58 \\
Llama-8B & 0.56 & 0.59 \\
Gemma-12B & 0.68 & 0.64 \\
Llama-70B & 0.74 & 0.65 \\
DeepSeek-70B & 0.72 & 0.73 \\
Qwen-72B & 0.72 & 0.69 \\
Qwen-235B & 0.74 & 0.68 \\
Llama-400B & \textbf{0.77} & 0.73 \\
DeepSeek-V3.1 & 0.74 & 0.56 \\
GPT-4o & 0.71 & 0.60 \\
Claude-3.5-Sonnet & 0.71 & 0.71 \\
\hline
\ourapproach{} (Mt-7B, $P_m + E_m + V_{m'}$) & \multicolumn{2}{c|}{0.60} \\
\ourapproach{} (Ll-8B, $P_m + E_m + V_{m'}$) & \multicolumn{2}{c|}{0.66} \\
\ourapproach{} (Qw-72B, $P_m + E_m + V_m$) & \multicolumn{2}{c|}{0.71} \\
\ourapproach{} (Qw-235B, $P_m + E_m + V_m$) & \multicolumn{2}{c|}{\underline{0.76}} \\
\hline
\end{tabular}
\caption{Macro F1 on SciTab-OD dataset.}
\label{tab:scitab_od_baselines}
\end{table}
\noindent\textbf{SciTab-OD Dataset.} Table ~\ref{tab:scitab_od_baselines} compares \ourapproach{} with baseline models on SciTab-OD. \rudra{Qw-235B ($P_m + E_m + V_m$) achieves 0.76 Macro F1, outperforming prominent models GPT-4o and Claude-3.5-Sonnet by 7.04\%. While the largest baseline Llama-400B reaches 0.77, \ourapproach{} achieves 98.70\% of this performance using only 235B parameters, demonstrating superior parameter efficiency. Ll-8B ($P_m + E_m + V_{m'}$) reaches 0.66, substantially outperforming comparably sized baselines (Llama-8B, Mistral-7B), and approaching the performance of larger Gemma-12B (0.68 w COT).} \rudra{Across both open-domain datasets, these results show that strategic agent coordination enables competitive performance while handling hybrid content of tables and paragraphs effectively. }

\subsubsection{Error Analysis}
\label{error-analysis-for-open-domain-datasets}
Error analysis across open-domain datasets reveals three misinterpretation patterns when using an independent verifier: (1) over-strict compound claim, (2) inexact numerical matches 
and (3) comparative terms like "outperform" requiring strict dominance. Model-specific patterns emerge: small models (Mt-7B) struggle with retrieval-induced ambiguity, while large models (Qw-235B) benefit from same-model verifiers that maintain reasoning consistency. In contrast, independent GPT-20B verifiers lack domain knowledge present in large base models, causing verification errors. This explains why same-model configurations outperform independent verifiers for large models (details in Appendix~\ref{appendix:error-analysis-open-domain}).

\section{Conclusion \& Future Work}

We introduce \ourapproach{}, a multi-agent framework for claim verification from tabular data documents with three specialized agents---Planner, Executor, and Verifier. Using only prompt-based reasoning, \ourapproach{} achieves SOTA performance on SciTab and SemTab and matches the best performance on two other datasets, while enabling smaller configurations to approach best-model performance with substantial memory savings. \eat{Smaller configurations (27-92B parameters) achieve 80-96\% of best performance while consuming only 11.5-39.1\% of the memory, democratizing access to high-quality verification.} Future work includes incorporating specialized agents for mathematical reasoning and implementing feedback loops between NEI predictions and retrieval.


\section{Limitations}
\rudra{While MACE demonstrates strong performance across diverse model sizes and datasets, the multi-agent coordination introduces computational overhead, with runtime increasing by 5.3$\times$ to 27.5$\times$ depending on model size and verification setting. This tradeoff between accuracy and execution time may limit deployment in latency-sensitive applications, though it remains justified for scenarios prioritizing verification accuracy over response speed. Additionally, our work does not focus on specialized techniques for handling long tables or context window limits. While we successfully process tables of varying sizes across our evaluation datasets, extremely large tables that exceed model context windows would require preprocessing strategies such as table decomposition, summarization, or retrieval-based chunking, which we leave for future work.}
\bibliography{custom}
\bibliographystyle{acl_natbib}

\appendix

\section{Baseline Details} \label{appendix:baseline-details}
\squishlist
    \item PASTA \cite{gu2022pasta}: A novel framework pretraining DeBERTaV3 on 1.2 million WikiTables sentence-table pairs, achieving SOTA performance with 85.6\% on TabFact, surpassing prior models by 4.7 points and narrowing the human performance gap to 1.5 points (92.1\%).
    \item ProTrix \cite{Wu2024ProTrixBM}: Proposes a Plan-then-Reason framework that plans reasoning paths and assigns steps to program-based or textual reasoning for answering user queries over tables with context.
    \item TART \cite{lu2024tart}: Introduces a tool-augmented reasoning framework with a table formatter, tool maker, and explanation generator to enhance LLM performance and explainability on tables.
    \item Volta \cite{gautam2021volta}: A BERT-based model for table understanding, verifying statements and identifying evidence using transfer learning and standardized table formats.
    \item LKA \cite{zhao-yang-2022-table}: A self-labeled keypoint alignment approach enhancing table-based fact verification by exploring statement-table correlations through dual-view alignment and an MOE block.
    \item Tapas \cite{muller-etal-2021-tapas}: Adapts the binary TAPAS model for SemEval-2021 Task 9, employing artificial neutral examples and pre-training on counter-factual data and TABFACT to classify statements as entailed or refuted.
    \item Tapex \cite{liu2021tapex}: Proposes table pre-training via a synthetic SQL executor, achieving accuracy improvements of 23--48\% on WikiSQL, WikiTableQuestions, SQA, and TabFact, setting new state-of-the-art results.
    \item DeBERTAV3 \cite{he2006deberta}: Enhances DeBERTa with replaced token detection pre-training to jointly encode sentences and tables.
    \item Various models: Include open-source (Alpaca-7B, Vicuna-13B, LLama-13B) and closed-source (InstructGPT, GPT-4) models for table-related tasks.

\squishend

\begin{figure*}[t]
    \centering
    \begin{tikzpicture}
        \node [rectangle, draw=black, inner sep=5pt, text width=\textwidth, align=left] {
            \textbf{Action Plan:}\par
            \vspace{0.2cm}
            1. \textbf{Identify Relevant Data Rows:}\par
            \hspace*{1em} - Extract the "Water reused/recycled" row for "All Operations" across all years (2023-2017). (\itb{rows 1 and 5} of Table \ref{tab:MineTabFact_table})\par
            \hspace*{1em} - Extract the "Operational water use(4)" row for "All Operations" across all years (2023-2017). (\itb{Rows 1 and 6} of Table \ref{tab:MineTabFact_table})\par
            \vspace{0.2cm}
            2. \textbf{Calculate Reused/Recycled Water Percentage for Each Year:}\par
            \hspace*{1em} - For each year, compute the percentage of reused/recycled water relative to operational water use:\par
            \hspace*{2em} \[
            \text{Percentage} = \left(\frac{\text{Water reused/recycled}}{\text{Operational water use}}\right) \times 100
            \]\par
            \hspace*{1em} - Perform this calculation for all years (2023-2017).\par
            \vspace{0.2cm}
            3. \textbf{Aggregate Across All Years:}\par
            \hspace*{1em} - Sum the "Water reused/recycled" values across all years.\par
            \hspace*{1em} - Sum the "Operational water use" values across all years.\par
            \hspace*{1em} - Compute the overall percentage:\par
            \hspace*{2em} \[
            \text{Overall Percentage} = \left(\frac{\text{Total Reused/Recycled}}{\text{Total Operational Water Use}}\right) \times 100
            \]\par
            \vspace{0.2cm}
            4. \textbf{Compare with Claimed Percentage (55.82\%):}\par
            \hspace*{1em} - Compare the computed overall percentage with the claimed 55.82\%.\par
            \hspace*{1em} - If the computed percentage matches (within reasonable rounding), the claim is supported. Otherwise, it is refuted.\par
            \vspace{0.2cm}
            5. \textbf{Assumptions and Clarifications:}\par
            \hspace*{1em} - The claim refers to "all operational water use across all years," so we focus on the "All Operations" section, not "Mining Operations." (\itb{Rows 1 and 7} of Table \ref{tab:MineTabFact_table})\par
            \hspace*{1em} - The caption clarifies that "Operational water use(4)" is the sum of water withdrawals and water reused/recycled, which aligns with the claim's context. (\itb{caption of Table \ref{tab:MineTabFact_table}})\par
            \hspace*{1em} - The term "across all years" is interpreted as an aggregate of all years (2017--2023), not an average of yearly percentages.\par
            \vspace{0.2cm}
            6. \textbf{Potential Issues:}\par
            \hspace*{1em} - If the computed percentage deviates significantly from 55.82\%, the claim is refuted.\par
            \hspace*{1em} - Ensure no misinterpretation of "All Operations" vs. "Mining Operations" data (the claim is about "all operations"). \par
            \textbf{Steps for the Executor:} These steps are provided by the Planner for the Executor to perform. 
            
            1. Extract "Water reused/recycled" and "Operational water use(4)" for "All Operations" (2023--2017) (\itb{rows 5 and 6} of Table \ref{tab:MineTabFact_table}).\\
            2. Compute yearly percentages and the overall percentage.\\
            3. Compare the overall percentage to 55.82\%.\\
            4. Return "support" if the computed percentage is 55.82\% (or very close, e.g., rounding to two decimal places), otherwise "refute." \par
            \textbf{Final Verdict:}\\
            The Executor will compute and return either "support" or "refute" based on the above steps. (Note: The Planner does not compute the final percentage; this is the Executor's task.) 
        };
    \end{tikzpicture}
    \caption{Action plan, generated by \ourapproach{}'s Planner agent for verifying the claim \textit{`Reused/recycled water made up 55.82
    \% of all operational water use across all years'} using data from Table~\ref{tab:MineTabFact_table}.}
    \label{fig:action_plan from planner agent}
\end{figure*}

\begin{figure*}[t]
    \centering
    \begin{tikzpicture}
        \node [rectangle, draw=black, inner sep=5pt, text width=\textwidth, align=left] {
            \textbf{Step 1}: Extract `Water reused/recycled' for All Operations (2017-2023):- 2017: 176563 ML- 2018: 174688 ML- 2019: 148914 ML- 2020: 157641 ML- 2021: 138812 ML- 2022: 134131 ML- 2023: 168358 ML. (\itb{row 5} of Table \ref{tab:MineTabFact_table})\\ Total Reused/Recycled = 176563 + 174688 + 148914 + 157641 + 138812 + 134131 + 168358 = 1,099,107 ML \par
            \textbf{Step 2}: Extract `Operational water use(4)' for All Operations (2017-2023):- 2017: 291931 ML- 2018: 302835 ML- 2019: 275931 ML- 2020: 275925 ML- 2021: 256074 ML- 2022: 250449 ML- 2023: 315784 ML (\itb{row 6} of Table \ref{tab:MineTabFact_table})\\Total Operational Water Use = 291931 + 302835 + 275931 + 275925 + 256074 + 250449 + 315784 = 1,968,929 ML \par
            \textbf{Step 3}: Compute overall percentage:Overall Percentage = $(1,099,107 / 1,968,929) * 100 \approx 55.82\%$ \par
            \textbf{Step 4}: Compare with claimed percentage (55.82\%):The computed percentage (55.82\%) matches the claimed percentage exactly. \par
            \textbf{Step 5}: Assumptions:- The claim refers to `all operations', so we ignored `Mining Operations' data.- `Across all years' was interpreted as an aggregate sum, not yearly averages.- The caption confirms `Operational water use(4)' is the correct metric for the denominator.
        };
    \end{tikzpicture}
    \caption{Execution trace, generated by \ourapproach{}'s Executor agent for verifying the claim \textit{`Reused/recycled water made up 55.82
    \% of all operational water use across all years'} using data from Table~\ref{tab:MineTabFact_table}.}
    \label{fig:action_plan from executor agent}
\end{figure*}

\begin{figure*}[t]
\centering
\small
\begin{mdframed}[linewidth=1pt]
\textbf{System Message}: You are the Planner Agent. Your role is to carefully analyze the scientific claim, the provided table, and the table caption. You must design a clear plan for how the Executor Agent should verify the claim using the table.

\textbf{Important Guidelines:}
\begin{enumerate}[leftmargin=*, itemsep=2pt]
    \item \textbf{Negation Awareness}: Pay very close attention to negation words such as ``not'', ``no'', ``without'', ``fail to'', ``does not'', or ``cannot''. These drastically change the meaning of the claim and must be reflected in your plan.
    \item \textbf{Relevance}: Focus only on the information in the table and caption. Do not invent data.
    \item \textbf{Consistency Rule}: Your plan must lead to a verdict that is consistent with your explanation.
    \item \textbf{Clarity}: Write step-by-step instructions the Executor Agent can follow.
\end{enumerate}

\textbf{Response Format:}
\begin{verbatim}
<explanation>
Your reasoning here: identify the key claim components, which table 
columns/rows are relevant, and what needs to be compared or checked.
</explanation>
<plan>
Explicit, step-by-step instructions for the Executor Agent.
</plan>
\end{verbatim}

\textbf{Description}: Agent for generating a detailed action plan to process the table, caption, and headers. Initiates the analysis process.
\end{mdframed}
\caption{Overview of the Planner Agent's system message and description for closed-domain verification, outlining its role and guidelines for creating an action plan to verify claims from tabular data.}
\label{fig:planner_agent_overview_closed}
\end{figure*}

\begin{figure*}[t]
\centering
\small
\begin{mdframed}[linewidth=1pt]
\textbf{System Message}: You are the Executor Agent. Your role is to follow the Planner Agent's plan and determine whether the claim is supported, refuted, or unverifiable (``not enough info'').

\textbf{Meanings of Verdicts:}
\begin{itemize}[leftmargin=*, itemsep=2pt]
    \item \textbf{support} $\rightarrow$ The table provides clear, relevant evidence that directly backs up the claim in full.
    \item \textbf{refute} $\rightarrow$ The table provides clear, relevant evidence that directly contradicts the claim, OR for compound claims, if the main part of the claim is contradicted or unsupported.
    \item \textbf{not enough info} $\rightarrow$ The table does not provide the required data to verify the claim. Absence of relevant metrics, entities, or comparisons must always lead to ``not enough info,'' not ``refute.''
\end{itemize}

\textbf{Guidelines:}
\begin{enumerate}[leftmargin=*, itemsep=2pt]
    \item \textbf{Consistency Rule}: Your explanation and your verdict must not contradict each other.
    \item \textbf{Compound Claims}: Treat the main point of the claim as decisive. Auxiliary or unverifiable details should not force ``refute'' if the main claim is supported.
    \item \textbf{Directional \& Numeric Checks}: Pay close attention to directional words (e.g., ``increase,'' ``decrease'') and numeric trends. Accept approximate matches (e.g., ``about 5'' = 4.8--6.2).
    \item \textbf{Unclear Plans}: If the Planner's plan is vague or unexecutable, respond with ``revise'' instead of a verdict, explaining what needs clarification.
    \item \textbf{Evidential Fairness}: Terms like ``large margin'' or ``substantial'' should be judged relative to the scale of results, not by an arbitrary cutoff. Use ``refute'' only if the data shows systematic contradiction to the claim.
\end{enumerate}

\textbf{Response Format:}
\begin{verbatim}
<explanation>
Step-by-step reasoning here...
</explanation>
support/refute/not enough info OR "revise"
\end{verbatim}

\textbf{Description}: Executes the Planner agent's action plan exactly as written, in the same order, without creating new steps or disambiguating terms. For each step, labels it explicitly (Step X: $<$original text$>$) and explains execution inside a single $<$explanation$>$...$<$/explanation$>$ block. Ends with a verdict (``support'', ``refute'', or ``not enough info''). If the plan is unclear, responds with ``revise'' instead of a verdict and explains what needs revision.
\end{mdframed}
\caption{Overview of the Executor Agent's system message and description for closed-domain verification, outlining its role and guidelines for executing the verification plan and determining the final verdict.}
\label{fig:executor_agent_overview_closed}
\end{figure*}

\begin{figure*}[t]
\centering
\small
\begin{mdframed}[linewidth=1pt]
\textbf{System Message}: You are the Verifier Agent. Your role is to check the Executor Agent's explanation and verdict for correctness, consistency, and fairness.

\textbf{Meanings of Verdicts:}
\begin{itemize}[leftmargin=*, itemsep=2pt]
    \item \textbf{support} $\rightarrow$ The table provides clear, relevant evidence that directly backs up the claim in full.
    \item \textbf{refute} $\rightarrow$ The table provides clear, relevant evidence that directly contradicts the claim, OR for compound claims, if the main part of the claim is contradicted or unsupported.
    \item \textbf{not enough info} $\rightarrow$ The table does not provide the required data to verify the claim. Absence of relevant metrics, entities, or comparisons must always lead to ``not enough info,'' not ``refute.''
\end{itemize}

\textbf{Guidelines:}
\begin{enumerate}[leftmargin=*, itemsep=2pt]
    \item \textbf{Consistency Check}: The Executor's explanation must match the verdict.
    \item \textbf{Compound Claims}: Do not demand perfection. Auxiliary unverifiable clauses should not override a supported main claim.
    \item \textbf{Numeric \& Qualitative Terms}: Accept approximate matches (e.g., 4.8 $\approx$ ``5''). Interpret terms like ``large margin'' or ``substantial'' relative to the scale of the results, not as fixed thresholds.
    \item \textbf{Revision Trigger}: If the Executor's verdict does not align with these definitions and guidelines, respond with ``revise'' and explain what should be corrected.
    \item \textbf{Final Verdict}: If Executor's output is sound, repeat the verdict as-is. If flawed but fixable, return ``revise''. Do not invent a verdict yourself beyond support/refute/not enough info.
\end{enumerate}

\textbf{Response Format:}
\begin{verbatim}
<explanation>
Your validation reasoning here...
</explanation>
support/refute/not enough info OR "revise"
\end{verbatim}

\textbf{Description}: Critically audits the Executor's output for strict consistency with the evidence and claim wording. Approves the verdict only if the explanation is complete, precise, and logically sound. If there is any flaw, ambiguity, or overreach, responds with ``revise'' to request correction from the Executor.
\end{mdframed}
\caption{Overview of the Verifier Agent's system message and description for closed-domain verification, outlining its role and guidelines for validating the Executor's output and ensuring consistency before issuing the final verdict.}
\label{fig:verifier_agent_overview_closed}
\end{figure*}

\begin{figure*}[t]
\centering
\small
\begin{mdframed}[linewidth=1pt]
\textbf{System Message}: You are an Input Agent tasked with sending the claim, table, and caption to the Planner Agent. Stop communication after receiving a final verdict (``support'', ``refute'', or ``not enough info'') from the Verifier Agent.

\textbf{Description}: Agent for initiating the process by sending the claim, table, and caption to the Planner Agent.
\end{mdframed}
\caption{Overview of the User Agent's system message and description for closed-domain verification, outlining its role in initiating the verification process and managing communication flow.}
\label{fig:user_agent_overview_closed}
\end{figure*}


\begin{figure*}[t]
    \centering
    \begin{tikzpicture}
        \node [rectangle, draw=black, inner sep=5pt, text width=\textwidth, align=left] {
            \textbf{System Message}: You are the Planner Agent. Your role is to carefully analyze the claim along with the provided evidence, which may include multiple tables and/or multiple paragraphs. You must design a clear plan for how the Executor Agent should verify the claim using the available evidence.\par
            \vspace{0.2cm}
            {\small
            \textbf{Important Guidelines:}\par
            1. \textbf{Negation Awareness}: Pay very close attention to negation words such as ``not'', ``no'', ``without'', ``fail to'', ``does not'', or ``cannot''. These drastically change the meaning of the claim and must be reflected in your plan.\par
            2. \textbf{Relevance}: Focus only on the information in the given tables and paragraphs. Do not invent data.\par
            3. \textbf{Consistency Rule}: Your plan must lead to a verdict that is consistent with your explanation.\par
            4. \textbf{Clarity}: Write step-by-step instructions the Executor Agent can follow.\par
            \vspace{0.2cm}
            \textbf{Response Format:}\par
            \texttt{<explanation>}\par
            Your reasoning here: identify the key claim components, which evidence sources (tables/paragraphs) are relevant, and what needs to be compared or checked.\par
            \texttt{</explanation>}\par
            \texttt{<plan>}\par
            Explicit, step-by-step instructions for the Executor Agent.\par
            \texttt{</plan>}
            }\par
            \vspace{0.2cm}
            \textbf{Description}: Agent for generating a detailed action plan to process the claim along with multiple tables and/or paragraphs. Initiates the analysis process.
        };
    \end{tikzpicture}
    \caption{Overview of the Planner Agent's system message and description for open-domain verification, outlining its role and guidelines for creating an action plan to verify claims from tabular data documents with retrieved evidence.}
    \label{fig:planner_agent_overview}
\end{figure*}

\begin{figure*}[t]
    \centering
    \begin{tikzpicture}
        \node [rectangle, draw=black, inner sep=5pt, text width=\textwidth, align=left] {
            \textbf{System Message}: You are the Executor Agent. Your role is to follow the Planner Agent's plan and determine whether the claim is supported or refuted.\par
            \vspace{0.2cm}
            {\small
            \textbf{Meanings of Verdicts:}\par
            - \textbf{support} $\rightarrow$ The evidence (tables/paragraphs) provides clear, relevant support that backs up the claim in full.\par
            - \textbf{refute} $\rightarrow$ The evidence clearly contradicts the claim, OR for compound claims, if any essential part of the claim is contradicted or unsupported.\par
            \vspace{0.2cm}
            \textbf{Guidelines:}\par
            1. \textbf{Consistency Rule}: Your explanation and your verdict must not contradict each other.\par
            \hspace{1em}- Explicit contradiction $\rightarrow$ ``refute''\par
            \hspace{1em}- Clear evidence alignment $\rightarrow$ ``support''\par
            2. \textbf{Compound Claims}: If any essential part of the claim is unsupported or contradicted, the correct outcome is ``refute.''\par
            3. \textbf{Directional \& Numeric Checks}:\par
            \hspace{1em}- Pay close attention to directional words (e.g., ``increase,'' ``decrease'') and numeric trends.\par
            \hspace{1em}- Accept approximate matches (e.g., ``about 5'' = 4.8--6.2).\par
            \hspace{1em}- Do not misinterpret negation or claim direction.\par
            4. \textbf{Unclear Plans}: If the Planner's plan is vague or unexecutable, respond with ``revise'' instead of a verdict, explaining what needs clarification.\par
            5. \textbf{Evidential Fairness}:\par
            \hspace{1em}- ``Outperform'' or ``improvement'' does not require being strictly higher in every metric.\par
            \hspace{1em}- Equal performance in some metrics with clear improvements in others still counts as support, as long as no consistent evidence contradicts the claim.\par
            \hspace{1em}- Minor ties or marginal differences do not automatically refute a claim.\par
            \hspace{1em}- Terms like ``large margin'' or ``substantial'' should be judged \textbf{relative to the scale of results}, not by an arbitrary cutoff.\par
            \hspace{1em}- Use ``refute'' only if the data shows systematic contradiction to the claim.\par
            \vspace{0.2cm}
            \textbf{Response Format:}\par
            \texttt{<explanation>}\par
            Step-by-step reasoning here...\par
            \texttt{</explanation>}\par
            support/refute OR ``revise''
            }\par
            \vspace{0.2cm}
            \textbf{Description}: Executes the Planner agent's action plan exactly as written, in the same order, without creating new steps or disambiguating terms. For each step, labels it explicitly (Step X: <original text>) and explains execution inside a single <explanation>...</explanation> block. Ends with a verdict (``support'' or ``refute''). If the plan is unclear, responds with ``revise'' instead of a verdict and explains what needs revision.
        };
    \end{tikzpicture}
    \caption{Overview of the Executor Agent's system message and description for open-domain verification, outlining its role and guidelines for executing the verification plan across multiple evidence sources and determining the final verdict.}
    \label{fig:executor_agent_overview}
\end{figure*}

\begin{figure*}[t]
    \centering
    \begin{tikzpicture}
        \node [rectangle, draw=black, inner sep=5pt, text width=\textwidth, align=left] {
            \textbf{System Message}: You are the Verifier Agent. Your role is to check the Executor Agent's explanation and verdict for correctness, consistency, and fairness.\par
            \vspace{0.2cm}
            {\small
            \textbf{Meanings of Verdicts:}\par
            - \textbf{support} $\rightarrow$ The evidence (tables/paragraphs) provides clear, relevant support that backs up the claim in full.\par
            - \textbf{refute} $\rightarrow$ The evidence clearly contradicts the claim, OR for compound claims, if any essential part of the claim is contradicted or unsupported.\par
            \vspace{0.2cm}
            \textbf{Guidelines:}\par
            1. \textbf{Consistency Check}: The Executor's explanation must match the verdict.\par
            \hspace{1em}- Explicit contradiction $\rightarrow$ ``refute''\par
            \hspace{1em}- Clear evidence alignment $\rightarrow$ ``support''\par
            2. \textbf{Compound Claims}: If any essential part of the claim is unsupported or contradicted, the correct outcome is ``refute.''\par
            3. \textbf{Numeric \& Qualitative Terms}:\par
            \hspace{1em}- Accept approximate matches (e.g., 4.8 $\approx$ ``5'').\par
            \hspace{1em}- Interpret terms like ``large margin'' or ``substantial'' relative to the scale of the results, not as fixed thresholds.\par
            4. \textbf{Revision Trigger}: If the Executor's verdict does not align with these definitions and guidelines, respond with ``revise'' and explain what should be corrected.\par
            5. \textbf{Final Verdict}:\par
            \hspace{1em}- If Executor's output is sound, repeat the verdict as-is.\par
            \hspace{1em}- If flawed but fixable, return ``revise''.\par
            \hspace{1em}- Do not invent a verdict yourself beyond support/refute.\par
            \vspace{0.2cm}
            \textbf{Response Format:}\par
            \texttt{<explanation>}\par
            Your validation reasoning here...\par
            \texttt{</explanation>}\par
            support/refute OR ``revise''
            }\par
            \vspace{0.2cm}
            \textbf{Description}: Critically audits the Executor's output for strict consistency with the evidence and claim wording. Approves the verdict only if the explanation is complete, precise, and logically sound. If there is any flaw, ambiguity, or overreach, responds with ``revise'' to request correction from the Executor.
        };
    \end{tikzpicture}
    \caption{Overview of the Verifier Agent's system message and description for open-domain verification, outlining its role and guidelines for validating the Executor's output and ensuring consistency before issuing the final verdict.}
    \label{fig:verifier_agent_overview}
\end{figure*}

\begin{figure*}[t]
    \centering
    \begin{tikzpicture}
        \node [rectangle, draw=black, inner sep=5pt, text width=\textwidth, align=left] {
            \textbf{System Message}: You are an Input Agent tasked with sending the claim, along with multiple tables and/or multiple paragraphs, to the Planner Agent. Stop communication after receiving a final verdict (``support'' or ``refute'') from the Verifier Agent.\par
            \vspace{0.2cm}
            \textbf{Description}: Agent for initiating the process by sending the claim, along with multiple tables and/or paragraphs, to the Planner Agent.
        };
    \end{tikzpicture}
    \caption{Overview of the User Agent's system message and description for open-domain verification, outlining its role in initiating the verification process and managing communication flow.}
    \label{fig:user_agent_overview}
\end{figure*}

\section{Error Analysis for Open-Domain Datasets}
\label{appendix:error-analysis-open-domain}

To understand the performance patterns observed in open-domain settings, we conducted a detailed error analysis across three experimental setups: (1) FinDVer-Testmini with Qwen-235B, comparing same-model \ourapproach{} ($P_m + E_m + V_m$) against \ourapproach{} with GPT-20B verifier ($P_m + E_m + V_{m'}$), (2) SciTab-OD with Mistral-7B, comparing w COT against \ourapproach{} with GPT-20B verifier ($P_m + E_m + V_{m'}$), and (3) SciTab-OD with Qwen-72B, comparing w COT against \ourapproach{} with GPT-20B verifier ($P_m + E_m + V_{m'}$). For each setup, we analyzed cases where the better-performing configuration produced correct predictions while the GPT-20B verifier variant failed.

A notable finding from setups (2) and (3) is that for both Mistral-7B and Qwen-72B on SciTab-OD, simple COT reasoning outperforms the complex multi-agent system with external GPT-20B verifier. This suggests that the added complexity of multi-agent coordination with an external verifier can introduce new failure modes that offset potential benefits, particularly when the verifier's reasoning style differs from the base model's capabilities.

\textbf{Common Error Patterns Across Model Sizes.} Our analysis reveals three error categories that consistently appear across all model sizes (small, medium, and large) when using the GPT-20B verifier:

\textbf{Over-strict Compound Claim Interpretation.} The GPT-20B verifier applies rigid standards when evaluating compound claims, failing to distinguish essential from contextual components. This pattern appears across all model sizes on both FinDVer and SciTab-OD. Consider the following claim from FinDVer: ``The company's international franchising revenue increased by 40.6\% in fiscal 2023 due to more stores in operation despite facing inflation and competitive threats impacting overall business operations, including increased store labor costs and intense competition from various sectors.'' Evidence confirms the 40.6\% revenue increase and mentions inflation, but lacks explicit discussion of ``intense competition from various sectors.'' The GPT-20B verifier issues ``refute'' because this contextual detail is unsupported, despite the core revenue claim being verified. In contrast, same-model verifiers (whether Qwen-235B, Qwen-72B, or even Mistral-7B with appropriate configuration) correctly identify the revenue metric as essential and the `competitive' context as supplementary, issuing a ``support'' verdict. This pattern is particularly problematic in financial domains where claims routinely include business context alongside factual data.

\textbf{Numerical Hyperprecision.} The external verifier demonstrates excessive strictness with numerical values across all datasets. In FinDVer-TM, when a claim states a facility was ``increased to \$750 million'' but evidence shows \$729.2 million (\$460M drawn + \$269.2M available), the GPT-20B verifier issues ``refute'' due to the \$20.8M discrepancy (2.8\% difference). Same-model verifiers exhibit greater tolerance, recognizing that financial reporting often involves rounding. For SciTab-OD with Mistral-7B, numerical precision errors manifest differently: when comparing performance metrics like 0.94 vs 0.93, the verifier treats these as strictly unequal rather than approximately matched, leading to incorrect ``refute'' verdicts for claims about model performance improvements.

\textbf{Misinterpretation of Comparative Terms.} When claims use terms like ``outperform'' or ``better than,'' the GPT-20B verifier applies overly strict interpretations across all model sizes. From the Scitab dataset, the statement ``Models using BoC outperform models using BoW and ASM'' is evaluated against a table showing BoC beats ASM on all metrics but ties with BoW on some SVM and ANN metrics while beating it on others. The GPT-20B verifier issues ``refute'' because BoC does not strictly dominate BoW on every metric. However, Mistral-7B with simple COT correctly interprets ``outperform'' as ``better overall,'' issuing ``support.'' Qwen-72B exhibits the same pattern, demonstrating this is a verifier-specific issue rather than a base model capability problem.

\textbf{Model-Specific Error Patterns.} Beyond common errors, we observe distinct patterns that vary by base model size:

\textbf{Inference Capability Gaps (Large Models Only).} For Qwen-235B on FinDVer, errors involve the verifier's inability to make reasonable domain-informed inferences. In claim ie-val-191 about ``.... the company is still facing substantial doubt about continuing as a going concern due to accumulated deficits and past operating losses.'' evidence discusses ``accumulated deficits,'' ``past operating losses,'' and ``need for additional financing'' without using the exact phrase ``going concern.'' The GPT-20B verifier marks this unsupported, while Qwen-235B's same-model verifier correctly infers the connection. This suggests that GPT-20B verifier may lack the domain knowledge that larger models possess, creating a capability mismatch when verifying outputs from sophisticated base models.

\textbf{Retrieval-Induced Ambiguity (Small Models).} For Mistral-7B on SciTab-OD, errors stem from the verifier's inability to handle ambiguous evidence retrieved from open-domain sources. When retrieved tables contain incomplete information or require cross-referencing multiple sections, the GPT-20B verifier defaults to ``refute'' if any claim component lacks explicit evidence. Mistral-7B with COT, however, can work within these constraints by focusing on verifiable core assertions and noting limitations. This pattern does not appear for larger models (Qwen-72B, Qwen-235B), which better handle ambiguous retrieval results even with the external verifier.

\textbf{Implications.} These error patterns explain the performance gaps between same-model and GPT-20B verifier configurations across open-domain datasets. The common errors (compound claims, numerical precision, comparative terms) affect all model sizes, while model-specific patterns (inference gaps for large models, retrieval ambiguity for small models) suggest verifier selection should adapt to base model capabilities. For large models like Qwen-235B, maintaining reasoning consistency within the same model family outweighs potential benefits from external verification. For small models like Mistral-7B, an external verifier can help with complex reasoning but may introduce new errors when evidence is ambiguous. The finding that simple COT outperforms multi-agent systems with external verifiers for Mistral-7B and Qwen-72B on SciTab-OD further underscores that added coordination complexity can be detrimental when verifier capabilities do not align well with base model reasoning patterns.

\end{document}